\documentclass{article}
\usepackage{url}
\usepackage{doi}
\usepackage{hyperref}
\usepackage{fullpage}
\usepackage{subfigure}
\usepackage{amsmath}
\usepackage{amsthm}
\usepackage{amsfonts} 
\usepackage{amssymb}
\usepackage{verbatim}
\usepackage{enumerate}
\usepackage[acronym=true]{glossaries} 
\usepackage{algorithm}
\usepackage[noend]{algpseudocode}
\usepackage[table]{xcolor}
\usepackage{subfigure}
\usepackage{xspace}
\usepackage{booktabs}
\usepackage{microtype}
\usepackage{natbib}
\usepackage{graphicx}

\hypersetup{
    colorlinks,
    linkcolor={red!50!black},
    citecolor={blue!50!black},
    urlcolor={blue!80!black}
}

\newcommand{\Ord}{\mathcal{O}}
\newcommand{\Orlib}{{\sc or-library}\xspace}

\newcommand{\CBGAe}{\gls{CBGA}$^d_\epsilon$\xspace}

\algnewcommand{\IfThen}[2]{% \IfThenElse{<if>}{<then>}{<else>}
  \State \algorithmicif\ #1\ \algorithmicthen\ #2}
\newacronym{GA}{GA}			{\textit{Genetic Algorithm}}
\newacronym{MKP}{MKP}	    {\textit{Multidimensional Knapsack Problem}}
\newacronym{CBGA}{CBGA}		{\textit{Chu \& Beasley Genetic Algorithm}}
\newcommand{\x}{\boldsymbol{x}}
\title{Randomized heuristic repair for large-scale multidimensional knapsack problem}

\author{\small Jean P. Martins \\
{\small Ericsson Research, ER, Brazil}\\
\texttt{\small jean.martins@ericsson.com}}
\begin{document}
\maketitle

%
%
%
%		Abstract
%
%
\begin{abstract}
The multidimensional knapsack problem (MKP) is an NP-hard combinatorial optimization problem whose solution consists of determining a subset of items of maximum total profit that does not violate capacity constraints. Due to its hardness, large-scale MKP instances are usually a target for metaheuristics, a context in which effective feasibility maintenance strategies are crucial. In 1998, Chu and Beasley proposed an effective heuristic repair that is still relevant for recent metaheuristics. However, due to its deterministic nature, the diversity of solutions such heuristic provides is not sufficient for long runs. As a result, the search ceases to find new solutions after a while. This paper proposes an efficiency-based randomization strategy for the heuristic repair that increases the variability of the repaired solutions, without deteriorating quality and improves the overall results. 
\end{abstract}
%
%
%
%		Keywords
%
%

\section{Introduction}

%
%					The Multidimensional Knapsack Problem
%

\glsreset{MKP}
The \gls{MKP} is a well-known strongly NP-Hard combinatorial optimization problem~\citep{freville2004, puchinger2010mkp}. To solve an instance of the \gls{MKP} one must choose, from a set of $n$ items, a subset that yields the maximum total profit. Every item chosen contributes with a profit $p_j > 0$ ($j=1,\dots,n$) but also consumes $w_{ij}$ from the  resources available $r_i>0$ ($i=1,\dots, m$). Therefore, a solution is \textit{feasible} only if the total of resources it consumes does not surpass the amount available. By representing solutions as $n$-dimensional binary vectors, the following integer programming model defines the \gls{MKP}:
\begin{alignat*}{3}
\max   \quad& f(\x) = \sum_{j=1}^{n} p_j x_j, &  \\
\text{subject to} \quad& \sum_{j=1}^{n} w_{ij} x_j \leq r_i &\quad i=1,\dots,m,\\
                      & x_j\in\{0,1\}, &\quad j=1,\dots,n.
\end{alignat*}

The \gls{MKP} is considerably harder than its uni-dimensional counterpart, not admitting an efficient polynomial-time approximation scheme even for $m=2$~\citep{Kulik2010}. Furthermore, \gls{MKP}'s hardness increases with $m$, and larger instances still cannot be efficiently solved to optimality~\citep{mansini2012}. Due to these limitations, metaheuristics have been the most successful alternatives to solve large instances (\Orlib\footnote{\texttt{http://people.brunel.ac.uk/~mastjjb/jeb/info.html}}). Hybrid methods, incorporating tabu-search, evolutionary algorithms, linear programming, and branch and bound techniques,  produced most of the best-known solutions for such instances~\citep{puchinger2010mkp}.

%One of the earliest references concerning \glspl{MKP} was a dynamic programming approach proposed by Gilmore and Gomory~\cite{gilmor1966}, which was further extended by Weingartner and Ness~\cite{weingartner1967} by embedding heuristics to the algorithm. Branch-and-bound was also used to improve dynamic programming approaches, as proposed by Marsten~\cite{marsten1977,marsten1978}. In the same context, Shih~\cite{shih1979branch} proposed the first branch-and-bound implementation based on linear programming. However, due to high space requirements and the limited computational capacity, such approaches were limited to very small problems. In order to circumvent such limitations, Gavish and Pirkul~\cite{gavish1985} developed a branch-and-bound using rules to reduce the problem size, obtaining better results than Shih's method. A more recent approach based on an approximated dynamic programming was proposed by Bertsimas~\cite{bertsimas2002}, while a hybridization with a branch-and-cut-procedure was proposed by Boyer~\cite{boyer2009}.

% Tabu-search methods, as developed by Glover and Kochenberger~\cite{glover1996} and further improved by Hanafi and Freville~\cite{hanafi1998} in 1998, were able to solve to optimality all the public instances available so far~\cite{freville2004}. 
 
\citet{chu1998} results can be considered a landmark regarding metaheuristics for the \gls{MKP}. The method they proposed, denoted as \gls{CBGA}, was one of the first metaheuristic to deal with large-scale \gls{MKP} instances, leading to several succeeding research studies~\cite{gottlieb2000b,gottlieb2001,tavares2006,tavares2008}. Additionally, the heuristic repair applied by \gls{CBGA} has been an effective alternative for feasibility maintenance commonly applied by evolutionary algorithms and related techniques. As a result, there have been many attempts to improve or replacing it~\cite{kong2008,wang2012,wang2012b,martins2013a,martins2013,chih2014,azad2014, Martins2020}, along with attempts to explain it \cite{Martins2014c, Martins2016, Martins2019}. %And last, the set of instances used to assess \gls{CBGA}'s performance became a standard benchmark for metaheuristics. 

This paper revisits \citet{chu1998}'s heuristic repair and tackles one of its weaknesses: determinism. As an attempt to overcome such limitation, we propose a randomized heuristic repair and compare its performance against \gls{CBGA}'s results. Section~\ref{sec:background}, reviews the main concepts needed for defining the type of heuristic repair discussed in the paper. Section~\ref{sec:randomization} describes how to enable an effective randomization. Sections~\ref{sec:methodology}, \ref{sec:experiments} and \ref{sec:results} define the implementation of the proposed heuristic repair on top of \gls{CBGA} along with the analysis of the results. Section~\ref{sec:conclusions} concludes the paper.

%\textcolor{red}{Several of \gls{CBGA} results have been improved or matched by many algorithms. Amongst them we mention \cite{vasquez2005} with an hybrid tabu-search/linear programming, \cite{mansini2012} with a core-based branch-and-bound and recently \cite{LAI2018282} with an hybrid of tabu-search and evolutionary algorithm; all of which have improved the best known solutions for some instances of the \Orlib.}

%Obviously, many other metaheuristics have been proposed for the \gls{MKP}~\cite{li2005,li2005b,li2012c}. 
%From now on, we use the \gls{CBGA} as reference framework, for a more complete review of \gls{MKP}'s peculiarities see \cite{kellerer2004knapsack, freville2004,puchinger2010mkp,hanafi2011}, for state-of-the-art results we suggest~\cite{boussier2010, hill2012,dellaCroce2012}.

%, including 
%
%
%
%
%     The Core Concept and Efficiencies
%
%
%
\section{Background}
\label{sec:background}
Many metaheuristics produce during the search infeasible candidate solutions. For the \gls{MKP}, an infeasible solution consists of a subset of items whose resource consumption exceeds the available amount, i.e., they violate capacity constraints. Therefore, the only way to repair an infeasible solution is by removing some of the items it contains. However, when removing a particular item, the resources that now become available might be enough to allow the addition of another item. As a result, a heuristic repair for the \gls{MKP} usually consists of two phases: (1) DROP items, (2) ADD items.

Naturally, by removing items of high value and low resource consumption in exchange for items of low value and high resource consumption, will yield a solution of low total profit. Therefore, a useful heuristic repair must consider removing and inserting items guided by an ordering that favors highly valuable solutions. Algorithm~\ref{alg:repair} formalizes these ideas: consider a candidate solution $\x$ and an ordering $\Ord$. The ordering indicates how to compare items in terms of the total value of the solutions they compose. Assume a non-increasing order of value for items, i.e., $\Ord_i$ is probably better than $\Ord_j$ if $i < j$.
\begin{algorithm}
	\caption{ {\sc heuristic-repair($\x$, $\Ord$)}}
	\label{alg:repair}
	\begin{algorithmic}[1]
		\ForAll {$j= \Ord_n,\dots,\Ord_1$} \Comment{\texttt{DROP items:}}
			\IfThen {$\x$ is feasible} {\textbf{break}}
			\State Remove $j$ from the knapsack%\leftarrow 0$
		\EndFor
		\ForAll {$j= \Ord_1,\dots,\Ord_n$} \Comment{\texttt{ADD items:}}
			   \If {$j$ fits in the knapsack}			   
			   	\State Add $j$ to the knapsack% \leftarrow 1$ \Comment{\texttt{// Inserts item $j$ in the knapsack}}
			   \EndIf
		\EndFor						
	\end{algorithmic}
\end{algorithm}

Algorithm~\ref{alg:repair} will produce high-quality solutions only if the ordering matches the \textit{efficiency} of the items. For example, in the uni-dimensional knapsack problem, every item $j$ has a profit of $p_j$ and consumes $w_j$ from the resource available. Therefore, by considering the items in non-increasing order of efficiency $e_j=p_j/w_j$, we would have an ordering that favors valuable items of small dimensions during the heuristic repair. Efficiency measures are also desirable for the multidimensional case. However, due to the multiple resources $w_{ji}$ involved, it is not straightforward to define a denominator for the equation. The usual approach consists of a weighted sum of the resources consumed by every item, as defined by Equation~\eqref{eq:edual}.
\begin{align}
e_j=\dfrac{p_j}{\sum_{i=1}^m \lambda_i \cdot w_{ij}},\forall j=1,\dots,n.\label{eq:edual}
\end{align}

From a comprehensive set of experimental, \citet{puchinger2010mkp} argued that the most effective weights $\lambda_i$, would be optimal solutions for the dual linear programming relaxation of the \gls{MKP}. Indeed, that was the weight vector employed by \citet{chu1998} in their experiments. We denote as $e_j^\text{dual}$ the efficiencies computed by the use of such weight vectors in equation~\eqref{eq:edual}. Additionally, we denote as  $\Ord^\text{dual}$ the ordering of the knapsack items that follow from such efficiencies.

Efficiencies provide reasonable estimates of how likely every item belongs to optimal solutions. Therefore, if an item has a high value of efficiency, it will probably be present in an optimal solution. On the other hand, if it has low value, it will most likely be rejected. Unfortunately, as the number of constraints grows ($m$),  the efficiency values for many items become too close to discriminate them (named as the \textit{core items}). As a result, for large \gls{MKP} instances, the ordering $\Ord^\text{dual}$ is less informative for the heuristic repair which hinders its effectiveness. As shown by~\citet{martins2014}, that seems to be the case for the \gls{CBGA}, with the algorithm struggling to decide if \textit{core items} should be put in the knapsack even in very long runs.

\section{Efficiency groups and randomization}
\label{sec:randomization}

The hypothesis we discuss from now on is that by exploring the space of orderings along with the search for better solutions, there might be an increase in the chances of finding better solutions. The question then would be how to perform such exploration without destroying the valuable information provided by the efficiencies. For that, we introduce the notion of \textit{efficiency groups}.

Assume, for the sake of argument, that we choose two items randomly and swap their positions in the ordering. If the efficiency of one of the items is much higher than the other, such an arbitrary exchange in the order is likely to deteriorate the quality of the solutions produced by the heuristic repair. On the other hand, if their efficiency values are close,  we expect the impact on the heuristic repair to be less disruptive. 

\begin{figure}[h!]
	\centering
	\subfigure[Original efficiencies]{
	\includegraphics[width=0.45\textwidth]{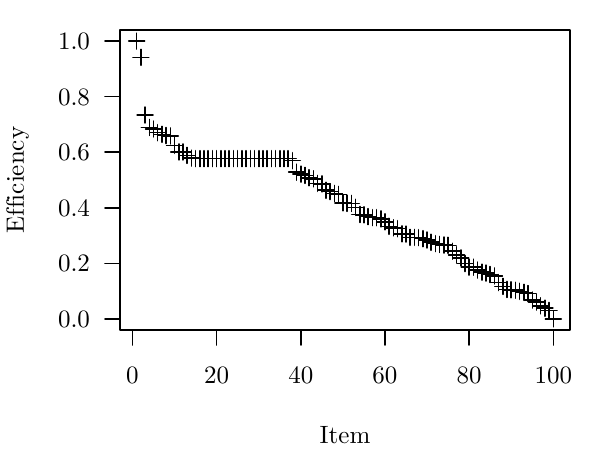}
	\label{fig:originaleff}
	}
	\subfigure[Efficiencies rounded to the first decimal case.]{
	\includegraphics[width=0.45\textwidth]{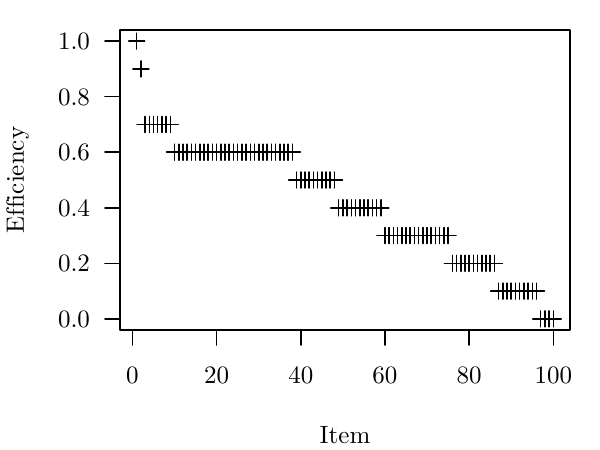}
	\label{fig:roundedeff}
	}
	\caption{Efficiency groups produced by rounding the original efficiencies to the first decimal case.}
\end{figure}

Observe for instance the Figure~\ref{fig:originaleff}, which shows the efficiencies\footnote{Scaled to the $[0,1]$ interval.} for the instance \texttt{30.100-00} (from \Orlib, $n=100$ items and $m=30$ resources) in non-decreasing order. Aside from a few items with high efficiency-values on the left-hand side, there is a plateau in which all the items share roughly the same efficiency (\textit{core items}). Additionally, all the following items share close efficiency values. Since these values define an ordering for the items, it is reasonable to question the strict ordering of items whose efficiencies are too close.

Figure~\ref{fig:roundedeff} shows us the same efficiencies but now rounded to the first decimal case ($d=1$). Rounding makes evident other plateaus, that indicate groups of items whose efficiencies are very close, i.e., an \textit{efficiency group}. Therefore, by adjusting the number of decimal cases for rounding, we can roughly control the sizes of these groups. In summary, an \textit{efficiency group} is a subset of two or more items, whose efficiencies are equal after rounding. Given this fact, we expect that items in the same group can have their ordering randomly modified without massive deterioration of the heuristic repair effectiveness. In this sense, efficiency groups would enable a smooth exploration of the space of orderings during the search.

\section{Methodology}
\label{sec:methodology}

From now on, we evaluate the possible benefits of two simple randomizing operators that modify the ordering of intra-group items given an initial ordering $\Ord^\text{dual}$.

\begin{description}
\item[Random-group swap ({\sc rg-swap}).] Randomly chooses an efficiency group, swap the positions of two randomly chosen items in the group.
\item[Random-group shuffle ({\sc rg-shuffle}).] Randomly chooses an efficiency group, shuffles the positions of possibly all of its items.
\end{description}

To verify the effectiveness of these operators, we must first decide how frequently randomization should take place. Since the \gls{CBGA} is an effective algorithm to find high-quality solutions fast, the use of intensive randomization during the first generations would slow-down its progress. To avoid such a drawback, we associate the exploration of new orderings with the number of improving solutions produced by heuristic repair during a generation. Whenever an ordering ceases to produce improving solutions, randomization will take place. Next section, describes a modified version of the \gls{CBGA} that implements such a strategy.

%
%
%
%	The Chu & Beasley Genetic Algorithm
%
%
%
\subsection{Chu \& Beasley GA with efficiency groups}\label{sec:cbga}
\newcommand{\p}{\mathcal{P}}
The \gls{CBGA} uses a direct representation (binary strings) and standard genetic algorithms' operators, such as binary tournament selection, uniform crossover and random mutation of two bits. Its only problem specific operator is the heuristic repair previously defined~\cite{chu1998, goldberg1989}. All candidate solutions that \gls{CBGA} produces are a result of the steps defined by Algorithm~\ref{alg:cbga-newsolution}.  

\begin{algorithm}[h!t]
	\caption{{\tt CBGA-newsolution($\p$)}}
	\label{alg:cbga-newsolution}
	\begin{algorithmic}[1]
	   \State Select $\x^1\in\p$ by a binary tournament 
	  \State Select $\x^2\in\p$ by a binary tournament 
	   \State $\x\leftarrow$ \texttt{uniformCrossover($\x^1$, $\x^2$)}
	   \State \texttt{flipTwoRandomBits($\x$)}
		\State \Return $\x$.		
	\end{algorithmic}
\end{algorithm}

\gls{CBGA} is a steady-state algorithm, which means it generates and evaluates one solution at the time. However, to control the activation of the randomization of efficiency orderings, we must account for the number of improving solutions produced during a generation. Therefore, we adapted the \gls{CBGA} to make that possible, with a generation consisting of $N$ attempts to produce improving solutions (where $N$ is the population size). Every time \gls{CBGA} produces a solution $\x$ it applies the heuristic repair to it. The resulting solution is an improvement if it is both unique and better than the worst solution in $\p$ (see Algorithm~\ref{alg:cbga-generation}).

\begin{algorithm}[h!t]
	\caption{{\tt CBGA-generation}($\p, N, \Ord^\text{dual}$)}
	\label{alg:cbga-generation}
	\begin{algorithmic}[1]	
		\State $m \gets 0$
		\State improvements $\gets 0$
		\While{$m < N$}
			   \State $\x\gets$ {\tt CBGA-newsolution($\p$)}
			   \State {\tt heuristic-repair}($\x$, $\Ord^\text{dual}$)
			   \If {$\x$ is unique and better than the worst solution in $\p$} 
			   	\State $\x$ substitutes the worst solution in $\p$
			   	\State improvements $\gets$ improvements $+ 1$
			   \EndIf
			   \State $m \gets m + 1$
		\EndWhile
		\State \Return improvements.		
	\end{algorithmic}
\end{algorithm}

The adapted algorithm \CBGAe supports the randomization of efficiency groups, with $d$ referring to the number of decimal cases used for rounding the efficiencies and $\epsilon$ standing for \textit{efficiency} (see Algorithm~\ref{alg:cbga})\footnote{\url{https://gitlab.com/jeanpm/mkp-egroups}}. The first step in \CBGAe is to sort the indexes of the items in non-increasing order according to $e^\text{dual}$. Next step is to store the efficiency groups in $g$. Next step, generates a population of $N$ random candidate solutions. All candidate solutions undergo heuristic repair and the search starts. At the end of every generation, if no improving solutions were produced, the randomization of the ordering takes place (where {\tt rand-ordering} is a placeholder for {\sc rg-swap} or {\sc rg-shuffle}). 

\begin{algorithm}[h!t]
	\caption{\CBGAe}
	\label{alg:cbga}
	\begin{algorithmic}[1]
		\Require The population size $N$
		\State $\Ord^\text{dual}\leftarrow$ \texttt{sort($\{1,\dots,n\}$, $e^\text{dual}$)} 
		\State $g \gets $ {\tt get-efficiency-groups}($\Ord^\text{dual}, d$)
		\State Generates $N$ random candidate solutions in $\p$

		\State {\tt heuristic-repair}($\x$, $\Ord^\text{dual}$), $\forall \x\in\p$

		\While{Stop criteria not met}			
			   \State $m \gets $ {\tt CBGA-generation}($\p$, $N$, $\Ord^\text{dual}$)
			   \IfThen {$m \leq 0$} {{\tt rand-ordering}($\Ord^\text{dual}$, $g$)}
		\EndWhile
		\State \Return the best solution in $\p$.		
	\end{algorithmic}
\end{algorithm}

%
%
%
%			 					EXPERIMENTS
%
%
%
\section{Experiments}
\label{sec:experiments}

The instances provided by \citet{chu1998} have been widely used in the literature to benchmark methods for the \gls{MKP}, and it will also be the basis for our experiments. In this paper we use larger-scale instances provided by Glover, the set comprises instances of size $n\in\{100,150,200,500,1500,2500\}$, for each size there are instances of dimension $m\in\{15,25,50,100\}$. Not every pair $n-m$ is available, there are $11$ instances\footnote{\texttt{http://www.info.univ-angers.fr/pub/hao/mkp.html}}.

To evaluate how the rounding impacts the efficiency groups, we consider two options $d=1,2$. With $d=1$ the efficiency groups are large, enabling a more intense exploration of the ordering space. As we increase $d$, the efficiency groups shrink, restricting the exploration of the orderings.

Regarding the randomizing operators, there are also two options. Given a randomly chosen efficiency group, the {\sc rg-swap} operator will exchange the position of two items, leading to subtle modification of the previous ordering. The {\sc rg-shuffle}, on the other hand, will possibly exchange the position of all items in the group, leading to a drastic modification of the previous ordering. 

The combination of two operators with two rounding options $d=1,2$ yields four algorithms to be compared with the original \gls{CBGA}, which we abbreviate as follows:

\begin{enumerate}
    \item $Sw_{d=1}$: {\sc rg-swap} with $d=1$,
    \item $Sw_{d=2}$: {\sc rg-swap} with $d=2$,
    \item $Sh_{d=1}$: {\sc rg-shuffle} with $d=1$,
    \item $Sh_{d=2}$: {\sc rg-shuffle} with $d=2$,
    \item $CBGA$: original algorithm.
\end{enumerate}

We ran all algorithms until finding a solution of quality equivalent to the best known, or until performing $10^8$ evaluations of the objective function. The average results from $30$ runs, for every instance, is reported. The tables of results contain four main parts:
\begin{enumerate}
    \item Instance name, 
    \item Best known: indicates the objective value of the best known feasible solution in the literature, 
    \item Gap: group of columns indicating the average gap of the solutions found in relation to the best known; 
    \item Time: group of columns indicating the average running time (in seconds) of every algorithm. 
\end{enumerate}

Additionally, an star symbol next to the gap is used to indicate if the best known solution was found at least once during the runs. The last row in every table summarizes the results by giving the overall number of ``wins'' of every algorithm, regarding solution quality (smallest gap) and running time (fastest).

%
%
%
%							Results
%
%
%
\section{Results and Discussion}\label{sec:results}
This section describes the results achieved by all five algorithms defined in the previous section. The results are organized first by the number of resources (i.e., the number of dimensions $m$) and second by the number of items ($n$) of the \gls{MKP} instances. Every instance class is denoted by the prefix $m$-$n$, where $m\in\{5,10,30\}$ and $n\in\{100,250,500\}$. In all tables, a star symbol ($*$) indicates that the particular algorithm has found a solution of quality equivalent to the best known at least once. Additionally, the last row always accounts for the number of times each algorithm was the best overall.

% latex table generated in R 3.3.3 by xtable 1.8-4 package
% Wed May 22 16:42:53 2019
\begin{table}[ht]
\centering
\begingroup\scriptsize
\begin{tabular}{|lr|rrrrr|rrrrr|}
  \cline{3-12}\multicolumn{2}{c}{}& \multicolumn{5}{|c|}{\sc gap} & \multicolumn{5}{|c|}{\sc time(s)}\\ \toprule
{gk} & Best & $Sw_{d=1}$ & $Sw_{d=2}$ & $Sh_{d=1}$ & $Sh_{d=2}$ & $CBGA$ & $Sw_{d=1}$ & $Sw_{d=2}$ & $Sh_{d=1}$ & $Sh_{d=2}$ & $CBGA$ \\ 
  \midrule
01 & 3766 & *0 & *0 & *0 & *0 & *0 & 6.38 & 4.64 & \cellcolor{red!40}10.68 & \cellcolor{green!40}2.22 & 3.04 \\ 
  02 & 3958 & *0 & *0 & *0 & *0 & *0 & 0.86 & \cellcolor{red!40}0.9 & 0.76 & 0.84 & \cellcolor{green!40}0.74 \\ 
  03 & 5656 & \cellcolor{red!40}6 & \cellcolor{red!40}*6 & \cellcolor{red!40}*6 & \cellcolor{red!40}*6 & \cellcolor{green!40}*5 & 867.95 & 1043.76 & \cellcolor{red!40}1094.35 & \cellcolor{green!40}793.34 & 1055.79 \\ 
  04 & 5767 & \cellcolor{red!40}*4 & \cellcolor{green!40}*3 & \cellcolor{red!40}*4 & \cellcolor{green!40}*3 & \cellcolor{green!40}*3 & 1271.69 & 1381.87 & 1217.66 & \cellcolor{green!40}1197.24 & \cellcolor{red!40}1598.97 \\ 
  05 & 7561 & \cellcolor{red!40}5 & \cellcolor{green!40}2 & 4.5 & \cellcolor{green!40}2 & \cellcolor{green!40}2 & \cellcolor{red!40}1418.96 & 1412.38 & 1398.89 & \cellcolor{green!40}1051.67 & 1357.1 \\ 
  06 & 7680 & 9 & 7 & \cellcolor{red!40}10 & \cellcolor{green!40}6 & 7 & 2131.05 & \cellcolor{red!40}2151.51 & 1679.85 & 1685.75 & \cellcolor{green!40}1651.79 \\ 
  07 & 19220 & 10 & 8 & \cellcolor{red!40}11.5 & \cellcolor{green!40}6.5 & 7 & 2704.94 & 3639.42 & 2938.57 & \cellcolor{red!40}3654.83 & \cellcolor{green!40}2696.44 \\ 
  08 & 18806 & \cellcolor{red!40}19.5 & \cellcolor{green!40}11 & \cellcolor{red!40}19.5 & 12 & \cellcolor{green!40}11 & 4223.56 & 5641.86 & \cellcolor{red!40}5713.04 & 4542.04 & \cellcolor{green!40}4131.3 \\ 
  09 & 58087 & 78 & 50 & \cellcolor{red!40}81 & 50.5 & \cellcolor{green!40}46.5 & 9190.78 & 10775.29 & 9819.7 & \cellcolor{green!40}9052.76 & \cellcolor{red!40}12148.83 \\ 
  10 & 57295 & 110.5 & 93.5 & \cellcolor{red!40}124.5 & \cellcolor{green!40}90 & 95.5 & \cellcolor{red!40}22183.19 & 21824.2 & 22102.64 & 17187.49 & \cellcolor{green!40}16101.22 \\ 
  11 & 95237 & 321 & 268 & \cellcolor{red!40}341 & 278.5 & \cellcolor{green!40}265 & 61362.53 & \cellcolor{green!40}50609.56 & 61922.44 & \cellcolor{red!40}69931.74 & 52918.18 \\ 
   & Wins & 0 & 3 & 0 & 5 & 6 & 0 & 1 & 0 & 5 & 5 \\ 
   \bottomrule
\end{tabular}
\endgroup
\end{table}

\section{Conclusion}\label{sec:conclusions}
The use of \textit{efficiency} measures to estimate the quality of knapsack items is a common heuristic used to guide the search for solutions for the \gls{MKP}. Such estimates induce an ordering for the items, that can be used during the search to prefer one item instead of others. However, if the ordering employed is not accurate, it might bias the search far from regions in the search space with high-quality solutions. 

This paper evaluated how a search in the space of orderings could improve the search for solutions. For that, we proposed a randomization strategy that acts by modifying the items ordering, always that improving solutions cease to be found. Such a strategy relies on the concept of \textit{efficiency groups} to avoid deterioration of the heuristic information provided by the initial ordering. Four variants of our proposal were implemented and compared to the original \gls{CBGA} in 270 \Orlib \gls{MKP} instances. Those variants differ by the size of the \textit{efficiency groups} they induce (depends on the parameter $d$) and how intensely they modify the items' ordering (swap of two items or shuffling the items within a group).

The results were encouraging and all variants of the randomized heuristic repair led to improvements in relation to the \gls{CBGA}. Such improvements consist of considerably smaller running times (more than ten times faster in some cases), smaller average gap to the best-known solutions and ability of finding solutions equivalent to the best known in cases where the \gls{CBGA} has failed. 

As a drawback, the quality of the results seems to depend on the randomization strategy chosen to modify the efficiency groups and the individual instance characteristics. So far, we could not identify a pattern on when to choose an aggressive strategy ({\sc rg-shuffle}) or a less disruptive one ({\sc rg-swap}). Additionally, there were cases where the \gls{CBGA} still achieved the best results overall, which makes the analysis of results even more difficult. 

For future work, it would be interesting to understand how the rounding parameter $d$ and the randomization strategies interact on different problem instances in order to propose a more robust combination. Another possibility would be to investigate the usefulness of \textit{efficiency groups} for improving local search procedures and also verify if they could be employed to define instance-specific strategies for modifying the orderings.

%
%
%
%			Bibliography
%
\section*{References}
\bibliographystyle{plainnat}
\bibliography{main}

\begin{thebibliography}{22}
\providecommand{\natexlab}[1]{#1}
\providecommand{\url}[1]{\texttt{#1}}
\expandafter\ifx\csname urlstyle\endcsname\relax
  \providecommand{\doi}[1]{doi: #1}\else
  \providecommand{\doi}{doi: \begingroup \urlstyle{rm}\Url}\fi

\bibitem[Azad et~al.(2014)Azad, Rocha, and Fernandes]{azad2014}
Md. Abul~Kalam Azad, Ana Maria~A.C. Rocha, and Edite~M.G.P. Fernandes.
\newblock Improved binary artificial fish swarm algorithm for the 0--1
  multidimensional knapsack problems.
\newblock \emph{Swarm and Evolutionary Computation}, 14\penalty0 (0):\penalty0
  66--75, 2014.
\newblock \doi{10.1016/j.swevo.2013.09.002}.

\bibitem[Chih et~al.(2014)Chih, Lin, Chern, and Ou]{chih2014}
Mingchang Chih, Chin-Jung Lin, Maw-Sheng Chern, and Tsung-Yin Ou.
\newblock {Particle swarm optimization with time-varying acceleration
  coefficients for the multidimensional knapsack problem}.
\newblock \emph{Applied Mathematical Modelling}, 38\penalty0 (4):\penalty0
  1338--1350, 2014.
\newblock \doi{10.1016/j.apm.2013.08.009}.

\bibitem[Chu and Beasley(1998)]{chu1998}
P.C. Chu and J.E. Beasley.
\newblock {A Genetic Algorithm for the Multidimensional Knapsack Problem}.
\newblock \emph{Journal of Heuristics}, 4:\penalty0 63--86, 1998.
\newblock \doi{10.1023/A:1009642405419}.

\bibitem[Freville(2004)]{freville2004}
Arnaud Freville.
\newblock {The multidimensional 0--1 knapsack problem: An overview}.
\newblock \emph{European Journal of Operational Research}, 155\penalty0
  (1):\penalty0 1--21, 2004.
\newblock \doi{10.1016/S0377-2217(03)00274-1}.

\bibitem[Goldberg(1989)]{goldberg1989}
D.E. Goldberg.
\newblock {Genetic algorithms and Walsh functions: Part I, a gentle
  introduction}.
\newblock \emph{Complex Systems}, 3\penalty0 (2):\penalty0 129--152, 1989.

\bibitem[Gottlieb(2000)]{gottlieb2000b}
Jens Gottlieb.
\newblock {Permutation-based Evolutionary Algorithms for Multidimensional
  Knapsack Problems}.
\newblock In \emph{{Proceedings of the 2000 ACM Symposium on Applied Computing
  - Volume 1}}, {SAC '00}, pages 408--414. ACM, 2000.
\newblock \doi{10.1145/335603.335866}.

\bibitem[Gottlieb(2001)]{gottlieb2001}
Jens Gottlieb.
\newblock {On the Feasibility Problem of Penalty-Based Evolutionary Algorithms
  for Knapsack Problems}.
\newblock In \emph{{Applications of Evolutionary Computing}}, volume 2037 of
  \emph{{Lecture Notes in Computer Science}}, pages 50--59. Springer Berlin
  Heidelberg, 2001.
\newblock \doi{10.1007/3-540-45365-2_6}.

\bibitem[Kong et~al.(2008)Kong, Tian, and Kao]{kong2008}
Min Kong, Peng Tian, and Yucheng Kao.
\newblock {A new ant colony optimization algorithm for the multidimensional
  Knapsack problem}.
\newblock \emph{Computers \& Operations Research}, 35\penalty0 (8):\penalty0
  2672--2683, 2008.
\newblock \doi{10.1016/j.cor.2006.12.029}.

\bibitem[Kulik and Shachnai(2010)]{Kulik2010}
Ariel Kulik and Hadas Shachnai.
\newblock {There is no EPTAS for two-dimensional knapsack}.
\newblock \emph{Information Processing Letters}, 110\penalty0 (16):\penalty0
  707--710, 2010.
\newblock \doi{10.1016/j.ipl.2010.05.031}.

\bibitem[Mansini and Speranza(2012)]{mansini2012}
Renata Mansini and M.~Grazia Speranza.
\newblock Coral: An exact algorithm for the multidimensional knapsack problem.
\newblock \emph{INFORMS Journal on Computing}, 24\penalty0 (3):\penalty0
  399--415, 2012.
\newblock \doi{10.1287/ijoc.1110.0460}.

\bibitem[Martins and Delbem(2016)]{Martins2016}
Jean~P. Martins and Alexandre~C.B. Delbem.
\newblock Pairwise independence and its impact on estimation of distribution
  algorithms.
\newblock \emph{Swarm and Evolutionary Computation}, 27:\penalty0 80--96, apr
  2016.
\newblock \doi{10.1016/j.swevo.2015.10.001}.
\newblock URL \url{http://dx.doi.org/10.1016/j.swevo.2015.10.001}.

\bibitem[Martins and Delbem(2019)]{Martins2019}
Jean~P. Martins and Alexandre~C.B. Delbem.
\newblock Reproductive bias, linkage learning and diversity preservation in
  bi-objective evolutionary optimization.
\newblock \emph{Swarm and Evolutionary Computation}, 48:\penalty0 145--155, aug
  2019.
\newblock \doi{10.1016/j.swevo.2019.04.005}.
\newblock URL \url{https://doi.org/10.1016%2Fj.swevo.2019.04.005}.

\bibitem[Martins and Ribas(2020)]{Martins2020}
Jean~P. Martins and Bruno~C. Ribas.
\newblock A randomized heuristic repair for the multidimensional knapsack
  problem.
\newblock \emph{Optimization Letters}, 15\penalty0 (2):\penalty0 337--355, jun
  2020.
\newblock \doi{10.1007/s11590-020-01611-1}.
\newblock URL \url{https://doi.org/10.1007%2Fs11590-020-01611-1}.

\bibitem[Martins et~al.(2013)Martins, {Bringel Neto}, Crocomo, Vittori, and
  Delbem]{martins2013a}
Jean~P. Martins, Constancio {Bringel Neto}, Marcio~K. Crocomo, Karla Vittori,
  and Alexandre C.~B. Delbem.
\newblock {A Comparison of Linkage-learning-based Genetic Algorithms in
  Multidimensional knapsack Problems}.
\newblock In \emph{{IEEE Congress on Evolutionary Computation }}, volume~1 of
  \emph{CEC'2013}, pages 502--509, June 20-23 2013.
\newblock \doi{10.1109/CEC.2013.6557610}.

\bibitem[Martins et~al.(2014{\natexlab{a}})Martins, Longo, and
  Delbem]{martins2014}
Jean~P. Martins, Humberto Longo, and Alexandre~C.B. Delbem.
\newblock On the effectiveness of genetic algorithms for the multidimensional
  knapsack problem.
\newblock In \emph{Proceedings of the Companion of Genetic and Evolutionary
  Computation}, GECCO Comp '14, pages 73--74. ACM, 2014{\natexlab{a}}.
\newblock \doi{10.1145/2598394.2598477}.

\bibitem[Martins and Delbem(2013)]{martins2013}
Jean~Paulo Martins and Alexandre Claudio~Botazzo Delbem.
\newblock {The influence of linkage-learning in the linkage-tree GA when
  solving multidimensional knapsack problems}.
\newblock In \emph{{Proceeding of the conference on Genetic and Evolutionary
  Computation}}, {GECCO '13}, pages 821--828. ACM, 2013.
\newblock \doi{10.1145/2463372.2463476}.

\bibitem[Martins et~al.(2014{\natexlab{b}})Martins, Fonseca, and
  Delbem]{Martins2014c}
J.P. Martins, C.M. Fonseca, and A.C.B. Delbem.
\newblock On the performance of linkage-tree genetic algorithms for the
  multidimensional knapsack problem.
\newblock \emph{Neurocomputing}, 146:\penalty0 17--29, 2014{\natexlab{b}}.
\newblock \doi{10.1016/j.neucom.2014.04.069}.

\bibitem[Puchinger et~al.(2010)Puchinger, Raidl, and
  Pferschy]{puchinger2010mkp}
Jakob Puchinger, Günther~R Raidl, and Ulrich Pferschy.
\newblock {The multidimensional knapsack problem: Structure and algorithms}.
\newblock \emph{INFORMS Journal on Computing}, 22\penalty0 (2):\penalty0
  250--265, 2010.

\bibitem[Tavares et~al.(2006)Tavares, Pereira, and Costa]{tavares2006}
J.~Tavares, F.B. Pereira, and E.~Costa.
\newblock {The Role of Representation on the Multidimensional Knapsack Problem
  by means of Fitness Landscape Analysis}.
\newblock In \emph{{ IEEE Congress on Evolutionary Computation}}, CEC'2006,
  pages 2307--2314, 0-0 2006.
\newblock \doi{10.1109/CEC.2006.1688593}.

\bibitem[Tavares et~al.(2008)Tavares, Pereira, and Costa]{tavares2008}
J.~Tavares, F.B. Pereira, and E.~Costa.
\newblock {Multidimensional Knapsack Problem: A Fitness Landscape Analysis}.
\newblock \emph{IEEE Transactions on Systems, Man, and Cybernetics, Part B:
  Cybernetics}, 38\penalty0 (3):\penalty0 604--616, june 2008.
\newblock ISSN 1083-4419.
\newblock \doi{10.1109/TSMCB.2008.915539}.

\bibitem[Wang et~al.(2012{\natexlab{a}})Wang, Fu, Mao, Menhas, and
  Fei]{wang2012b}
Ling Wang, Xiping Fu, Yunfei Mao, Muhammad~Ilyas Menhas, and Minrui Fei.
\newblock {A novel modified binary differential evolution algorithm and its
  applications}.
\newblock \emph{Neurocomputing}, 98\penalty0 (0):\penalty0 55--75,
  2012{\natexlab{a}}.
\newblock \doi{10.1016/j.neucom.2011.11.033}.
\newblock Bio-inspired computing and applications (LSMS-ICSEE ' 2010).

\bibitem[Wang et~al.(2012{\natexlab{b}})Wang, Wang, and Xu]{wang2012}
Ling Wang, Sheng-yao Wang, and Ye~Xu.
\newblock {An effective hybrid EDA-based algorithm for solving multidimensional
  knapsack problem}.
\newblock \emph{Expert Systems with Applications}, 39\penalty0 (5):\penalty0
  5593--5599, 2012{\natexlab{b}}.
\newblock \doi{10.1016/j.eswa.2011.11.058}.

\end{thebibliography}

\end{document}